\documentclass[conference]{IEEEtran}
\IEEEoverridecommandlockouts

\usepackage{cite}
\usepackage{comment}
\usepackage{amsmath,amssymb,amsfonts}
\usepackage{algpseudocode}
\usepackage{algorithm}

\usepackage{graphicx}
\usepackage{textcomp}
\usepackage{xcolor}
\usepackage{multirow}
\usepackage{makecell}
\usepackage{hyperref}
\begin{document}

\title{Boosting Weak Positives for Text Based Person Search}


\author{
  \begin{minipage}{\textwidth}
    \begin{minipage}{0.245\textwidth}
      \centering
      Akshay Modi*
    \end{minipage}%
    \begin{minipage}{0.245\textwidth}
      \centering
      Ashhar Aziz*
    \end{minipage}%
    \begin{minipage}{0.245\textwidth}
      \centering
      Nilanjana Chatterjee
    \end{minipage}
    \begin{minipage}{0.245\textwidth}
      \centering
      A V Subramanyam
    \end{minipage}
    
    \vspace{1em}
    \begin{center}
      \texttt{\{akshay21166,ashhar21137,nilanjanac,subramanyam\}@iiitd.ac.in}
    \end{center}
    
  \end{minipage}
}

    

\maketitle

\begin{abstract}
Large vision-language models have revolutionized cross-modal object retrieval, but text-based person search (TBPS) remains a challenging task due to limited data and fine-grained nature of the task. Existing methods primarily focus on aligning image-text pairs into a common representation space, often disregarding the fact that real world positive image-text pairs share a varied degree of similarity in between them. This leads models  to prioritize easy pairs, and in some recent approaches, challenging samples are discarded as noise during training. In this work, we introduce a boosting technique that dynamically identifies and emphasizes these challenging samples during training. Our approach is motivated from classical boosting technique and dynamically updates the weights of the weak positives, wherein, the rank-1 match does not share the identity of the query. The weight allows these misranked pairs to contribute more towards the loss and the network has to pay more attention towards such samples. Our method achieves improved performance across four pedestrian datasets, demonstrating the effectiveness of our proposed module. Code is available \url{https://anonymous.4open.science/r/TBPS_Boosting}.
\end{abstract}
\renewcommand{\thefootnote}{}
\footnotetext{* denotes equal contribution}


\section{Introduction}
Text-Based Person Search (TBPS)\cite{b13} focuses on identifying a specific individual within a large image dataset using a free-form natural language description. This approach provides a practical solution for surveillance, especially in scenarios where a visual reference of the person of interest is unavailable. As a result, leveraging diverse natural language descriptions for person identification has emerged as a prominent research focus.

In recent years, vision-language pre-training models (VLMs) \cite{b12,b21}, have made remarkable advancements. These models are trained on extensive datasets to establish a unified semantic space and obtain an effective alignment between image and text embeddings. Depending on the training strategy, either new parameters are added and trained from scratch, or the large backbone networks are partially or fully fine-tuned to optimize performance.
Existing works based on VLMs adopt a standard contrastive loss framework to bring positive image-text pair embeddings closer while pushing negative embeddings apart within a shared semantic space \cite{b1},\cite{b2}. 

\begin{figure}[t]
    \centering
    \includegraphics[width=\columnwidth]{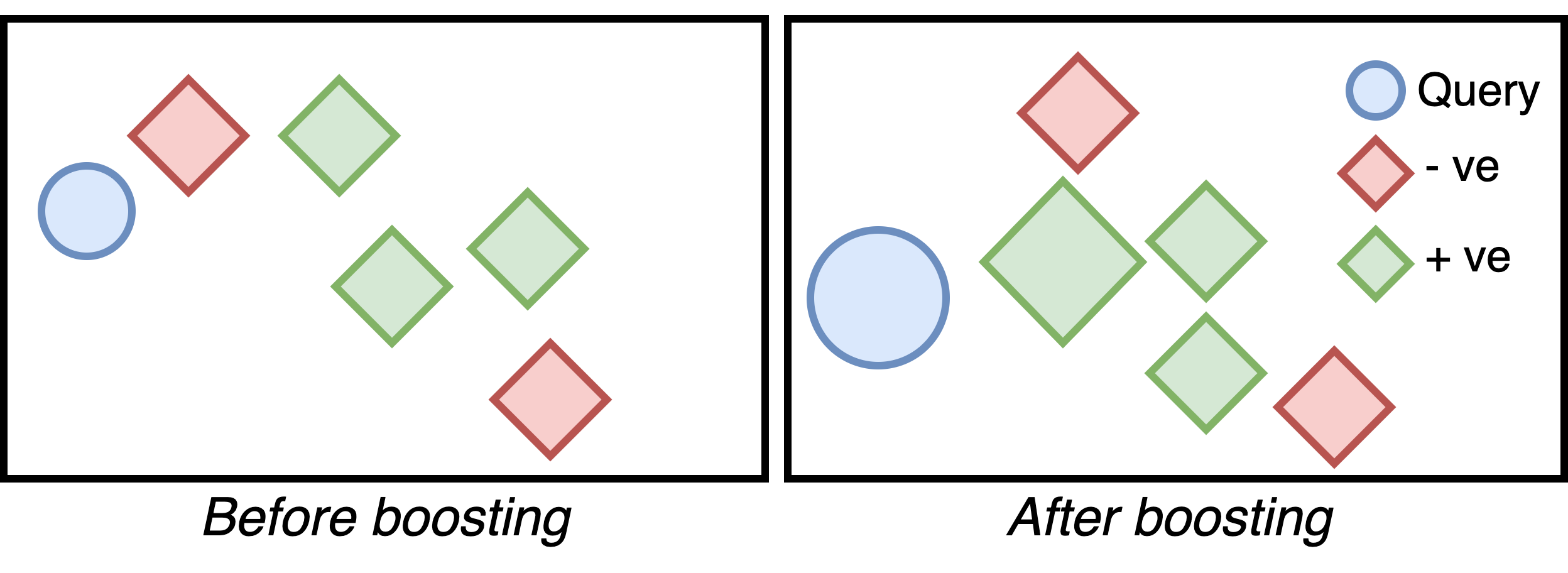}
    \caption{  Illustrative presentation of how positive and negative samples behave before and after our boosting method. Boosting increases weight of the positive pair which occurs at rank-$k$ ($k=2$ in this case). 
    After boosting, the negative sample is pushed away from the query while the second most similar positive sample is pulled. }
    \label{fig:intro}
\end{figure}
TBPS datasets inherently have high-variance due to the very nature of annotation process. The captions can be vague and the image quality may be suboptimal. Therefore, dynamic weighting or filtration processes have gained traction to manage noisy data points during model training.
RDE \cite{b16} addresses the noisy nature of data by introducing a noise correspondence process that dynamically weighs outlier image-text pairs. RDE, treats these outliers as noise and reduces their contribution during training. In contrast, our work identifies difficult samples or weak positive pairs  and demonstrates that dynamically boosting their weight leads to more effective model training. In  Figure \ref{fig:intro}, we demonstrate how boosting brings the true positives closer to each other by increasing their weight. Unlike methods that reduce or discard noisy data points to accelerate learning, our approach leverages the challenging samples to enhance model performance.

Our boosting mechanism draws inspiration from the classical AdaBoost \cite{b17} technique. Our contributions are as follows: 
\begin{itemize}[]
    \item We propose a novel approach for {selection and dynamic incorporation of rank-$k$ samples} for boosting. These samples are the $k^{th}$ most similar image (text) to the query text (image) that matches the ground truth but are not selected as the rank-1 prediction by the model. The significance of such rank-$k$ samples has been thoroughly tested through extensive experimentation.
    
    \item We design a {boosting-aware contrastive loss} function that integrates the boosted samples, allowing the model to emphasize learning from these difficult examples. This loss formulation enhances the model's robustness  by effectively balancing the learning process between general and challenging samples.
    \item Furthermore, we extend our boosting technique to various losses and demonstrate its effectiveness.
\end{itemize}

\section{Related Works}
Text-based person search was initially introduced by Li \textit{et al.} \cite{b13} with the release of the CUHK-PEDES benchmark dataset. Early research in this domain faced considerable difficulties in aligning image-text encodings generated by unimodal backbone networks \cite{b3} -\cite{b8}. These works can be categorized in terms of global \cite{b11}, and local \cite{b3},\cite{b18} feature matching.

Since the inception of large Vision-Language Models (VLMs) \cite{b12}, \cite{b21}, recent works \cite{b1}, \cite{b2}, \cite{b19} leverage VLM backbones for feature extraction and transfer learning. CFine \cite{b2} was the first to adopt CLIP \cite{b12} as its image-text feature backbone. Due to CLIP's lightweight architecture and efficiency, it has become a popular choice in TBPS, later adopted by IRRA\cite{b1} and RDE\cite{b19}. A recent work, APTM\cite{b22} has turned to pre-training it's backbone from scratch using a synthetic dataset consisting of only person image-caption pairs.  Even, if they have shown improved results, using general-purpose VLM like CLIP and pre-training from scratch still remain two distinct research methods.  RaSA\cite{b24} trained on ALBEF\cite{b21} shows improved performance over CLIP based TBPS models. But it is worth to note, all the pre-trained VLMs have varied number of training parameters making it difficult to make a direct comparison. TBPS-CLIP\cite{b22} has conducted extensive experiments on CLIP where they analyze multiple loss functions, training strategies and data augmentation processes. 

While VLMs have revolutionized cross-modal retrieval tasks, local network learning remains a key focus for mining the fine-grained information hidden within image patches. For instance, IRRA introduces an implicit relation learning local network that predicts caption words using image features. This is conceptually similar to the earlier work LapsScore\cite{b18}, which leverages colored images to predict the color of different attributes. On the other hand, RDE, utilizes local attention maps extracted from the backbone to enhance its local feature network.

However, despite the transition from unimodal backbones to VLMs, significant alignment with person-specific datasets is still required. Global feature matching entails extracting identity-specific embeddings and effectively aligning image-text pairs through contrastive losses. Image-text matching losses have been a central area of research in vision-text alignment, extending beyond TBPS. For example, IRRA incorporates ID loss\cite{b20} and designs the SDM loss, inspired by the CMPM loss\cite{b19}. Similarly, RDE introduces a triple alignment loss inspired by ranking loss. Among these, the InfoNCE\cite{b15} loss remains one of the most widely used and fundamental losses for training large-scale foundational VLMs like CLIP.

While significant research has focused on architecture design and alignment losses, less attention has been given to the nature of real-world person datasets. RDE highlights the discrepancies and varied nature of such data and designs a noise filtration system. In our work, we emphasize the lack of clean data in TBPS domain and real-world data inherently contains variance. Thus, it becomes crucial to handle weakly-aligned positive image-text pairs. To address this challenge, distinct from prior methods, we propose a novel boosting technique that enhances model robustness by effectively weighting challenging samples.


 \subsection{Preliminaries}
We briefly discuss the idea of boosting. In case of boosting, misclassified samples are boosted so that it contributes more to the loss. Let the weight of sample \(i\) at iteration \(n\) be \(w_i^n\). Then
the weights of the samples $x_i$ are updated as follows:
\[
w_i^{n+1} = w_i^n \cdot \exp\left(\alpha_n \cdot \mathbb{I}(h_n(x_i) \neq y_i)\right),
\]
where $h_n$ is a weak learner and $y_i$ is the label.
The loss can be represented as a function of $w_i^{n}$ and misclassified samples as $\mathcal{L}(w_i^{n}, \mathbb{I}(h_n(x_i) \neq y_i))$. On similar lines, we identify the pair of samples for which the gallery sample having the best similarity with the query does not have the same identity as that of query. We identify such samples and boost their respective weights.
\begin{figure*}[h]
    \centering
    \includegraphics[width=0.82\textwidth]{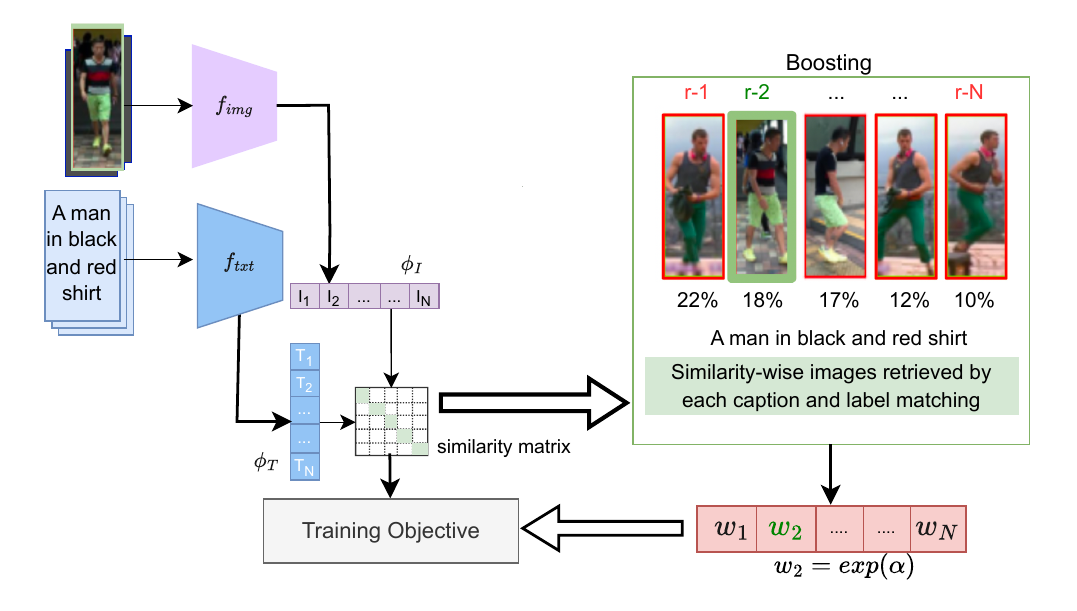}
    \caption{Images and captions are encoded using image and text encoders. The embeddings are used to calculate a similarity matrix, which interacts with the boosting module to dynamically identify and enhance weak positives. The similarity matrix and boosting weights are then passed to the training objective.}
    \label{fig:ema_diagram}
\end{figure*}
\section{Methodology}
In the ReID setting, we first identify misranked pairs.  Given a query, if the rank-1 retrieved sample does not have the same ID as that of query, then the misranked pair refers to retrieved sample at rank-$k$ which shares the same ID as that of query. We boost this misranked pair such that it contributes more to the loss and the network has to focus more on this pair. Eventually the correct match can move to rank-1. In the following, we explain the feature extraction and our boosting method.

\subsection{Feature extraction}
Given a dataset $\mathcal{D} = \{I_i, T_i, y_i\}_{i=1}^N$
where \( I_i \) represents the \( i \)-th image in the dataset, \( T_i \) is the corresponding text description for \( I_i \), and \( y_i \) is the class label or person identity label, and \( N \) is the total number of training samples.

For each image \( I_i \), we obtain its embedding from an image encoder, $f_{img}$, as $\phi_{{I}_{i}} = f_{{img}}(I_i)$, where \( \phi_{{I}} \in \mathbb{R}^d \). Similarly, for each text \( T_i \), the text encoder $f_{{txt}}$ generates the text embedding, $\phi_{{T}_{i}} = f_{{txt}}(T_i)$, where \( \phi_{{T}} \in \mathbb{R}^d \). In case of baseline CLIP model, we use the CLS token embeddings from the pretrained CLIP for extracting both image and text embeddings.

\subsection{Boosting}

We first establish a relation with AdaBoost using a single query, and a gallery set. 
Let \(q\) be the query, and \(g_1, g_2, \dots, g_K\) be the gallery samples ranked in descending order of similarity to \(q\). Suppose \(g_k\) is the correct match (positive) at rank-$k$, and \(g_1\) is the incorrect match (negative) at rank-1. 

We aim to emphasize the weights of the misranked positive pairs to push them to rank-1. Let the weight corresponding to a pair ($q, g_i$) be $w_i$. At the beginning of an epoch, we first initialize the weights $w_i$ to 1. We then define the weight update of a pair $\{(q, g_i)\}$ as:
\[
w_i \leftarrow w_i \cdot \exp\left(\alpha \cdot \mathbb{I}(y_{g_i} = y_q \cap y_{g_1} \neq y_q)\right),
\]
In our experiments, we set $exp(\alpha) = 1.6$. Thus we boost samples for which the query has a correct match at rank-$k$ and incorrect match at rank-1. For instance, the pair $(q,g_k)$ will recieve a weight update of $w_k = exp(\alpha)$ while all other pairs which do not share the label with $q$ will have weight of 1 itself. Unlike AdaBoost, we do not normalize the weights. In the following we explain the boosting when we have a set of query and gallery samples.

Let $T_i$ be the query text, $I_i$ be the corresponding image for $T_i$ at a  rank-$k$, where $k >$ 1. Let the image at rank-1 be $\bar{I}_i$. $y_{\bar{I}_i}$ be the label of $\bar{I}_i$. Now define the set $\mathcal{R}_k$ as:
\[
\mathcal{R}_k = \left\{ (T_i, I_i) \;\middle|\;
\begin{aligned}
& \text{rank}(T_i,I_i) = k, \\
& \text{rank}(T_i,\bar{I}_i) = 1, \\
& y_{\bar{I}_i} \neq y_i, \\
\end{aligned}
\right\}.
\]

The boosting coefficient \( w \) is assigned dynamically to emphasize the contribution of these rank-$k$ samples during training. Formally, \( w^k(i) \) corresponding to rank-$k$ is defined as:
\begin{equation}
w^k(i) = 
\begin{cases} 
exp(\alpha), & \text{if } (T_i, I_i) \in \mathcal{R}_k, \\
1, & \text{otherwise}.
\end{cases}
\end{equation}

This formulation ensures that the contribution of rank-$k$ samples is increased while maintaining natural contributions for all other samples. We now discuss how these weights are incorporated in the loss objective.

\subsection{Training Objective}
The boosted Image-Text Contrastive (ITC) loss dynamically adapts during training, ensuring that the model focuses on challenging samples while naturally learning from the unboosted samples. For text-to-image (t2i), the boosting aware ITC loss is defined as:

\begin{equation}
\mathcal{L}_{\text{t2i}} = -\frac{1}{B} \sum_{i=1}^B w_b^k(i)  \log \frac{\exp(\text{sim}(\phi_{I_i}, \phi_{T_i})/\tau)}{\sum_{j=1}^B \exp(\text{sim}(\phi_{I_j}, \phi_{T_i})/\tau)},
\label{eq:con_1}
\end{equation}
where $w_b^k$ is obtained from $w^k$ by finding a correspondence between the samples in the batch $B$ and the same pairs in $\mathcal{R}_k$. $\tau$ is the temperature hyperparameter.
In case of image-to-text (i2t), the loss is:

\begin{equation}
\mathcal{L}_{\text{i2t}} = -\frac{1}{B} \sum_{i=1}^B w_b^k(i)  \log \frac{\exp(\text{sim}(\phi_{I_i}, \phi_{T_i})/\tau)}{\sum_{j=1}^B \exp(\text{sim}(\phi_{I_i}, \phi_{T_j})/\tau)}. 
\label{eq:con_2}
\end{equation}

In Equations \ref{eq:con_1} and \ref{eq:con_2}, \( w_{b}^k(i) \) determines the coefficient for each positive pair. Thus, the weights of the samples which belong to the set $\mathcal{R}_k$ get boosted. As a consequence these samples contribute more to the loss and network focuses more on them to increase their rank.

Finally, the combined loss is calculated as:

\begin{equation}
\mathcal{L} = \frac{\mathcal{L}_{\text{i2t}} + \mathcal{L}_{\text{t2i}}}{2}.
\label{eq:total-loss}
\end{equation}

We use the CLIP-ViT-B/16 model fine-tuned with InfoNCE\cite{b15} loss as our Baseline. And we use equation  \ref{eq:total-loss} to train the boosted version CLIP+B.
\subsection{Extension to other loss functions}
To show the effectiveness of our module we combine our boosting technique with various losses. We study our boosting method with ID loss and SDM loss \cite{b1}. Given a batch of image-text pairs and their labels $\{I_i, T_i,y_i\}_{i=1}^B$, the loss is given as
\begin{equation}
    \mathcal{L}_{\text{ID}}(I,y) = -\sum_{i=1}^B y_i^{\top} \ln \hat{y}_i
\end{equation}
With our boosting module, this loss is modified as
\begin{equation}
    \mathcal{L}_{\text{ID+B}}(I,y) = -\sum_{i=1}^M w_b^{k}(i)y_i^{\top} \ln \hat{y}_i
\end{equation}
Similarly, we can compute it for captions. Boosted
SDM loss can be derived as 
\[
\mathcal{L}_{\text{SDM+B}} = KL(\mathbf{p}_i \| \mathbf{q}_i) = \sum_{i=1}^{B}\sum_{j=1}^{B} w_b^{k}(i)p_{i,j} \log \frac{p_{i,j}}{q_{i,j}}
\]
where $p_{i,j} = \frac{\exp\left(\text{sim}(\phi_{I_{i}}, \phi_{T_{i}}) / \tau \right)}{\sum_{n=1}^{B} \exp\left(\text{sim}(\phi_{I_{i}}, \phi_{T_{n}}) / \tau \right)} $ and $q_{i,j} = \frac{y_{i,j}}{\sum_{n=1}^{B} y_{i,n}}$
%
Both IRRA and RDE use CLIP as encoders with various loss functions for different objectives. This makes it possible for us to seamlessly combine our boosting method keeping the rest of the architecture same.
The updated objective for IRRA \cite{b1} combined with boosting is defined as:

\[
\mathcal{L}_{\text{IRRA+B}} = \mathcal{L}_{\text{IRR}} + \mathcal{L}_{\text{SDM+B}} + \mathcal{L}_{\text{ID+B}}.
\]
IRR module trains the local network and doesn't compute pairwise feature loss. So the loss $\mathcal{L}_{\text{IRR}}$ \cite{b1} is not boosted.
RDE \cite{b19}, on the other hand, originally utilizes only the TAL (Triple Alignment Loss) in its objective. We boost the TAL loss, similar to boosting the SDM loss, to train RDE.

\section{Experiments}
\subsection{Datasets and evaluation metrics}
We use four pedestrian datasets, CUHK-PEDES, ICFG-PEDES, RSTPReid, and IIITD-20K, for training and testing. We compare the results using rank-k (1,5,10) accuracy and mean average precision (mAP). We follow the evaluation protocol of IRRA \cite{b1}.

CUHK-PEDES\cite{b13} contains 11,003 identities, 34,054 images, and 68,108 textual descriptions. The validation and test sets include 3,078 and 3,074 images, along with 6,158 and 6,156 textual descriptions, respectively. Both sets have 1,000 unique identities.

ICFG-PEDES\cite{b6} consists of 54,522 images for 4,102 identities, with each image paired with a single textual description. The dataset is split into a training set and a test set. The training set contains 34,674 image-text pairs from 3,102 identities, while the test set includes 19,848 image-text pairs from the remaining 1,000 identities.

The RSTPReid\cite{b5} dataset contains 20,505 images representing 4,101 identities captured across 15 cameras. Each identity is associated with 5 images from different cameras, and each image is paired with 2 textual descriptions. The official data split divides the dataset into 3,701 identities for training, 200 for validation, and 200 for testing.

The IIITD-20K\cite{b14} dataset contains 20,000 images, each corresponding to a unique identity, with a total of 20,000 identities. Each image is annotated with 2 captions, and the average caption length is 35.9 words.

\subsection{Implementation details}
We use the CLIP ViT backbone as the image encoder and CLIP BERT as the text encoder. The input patch size is set to 
384$\times$128 and the hidden dimension is set to 512. The maximum context length for the text is set to 77 tokens. We train the model for 60 epochs. The initial learning rate is set to 1e-5. We tested our boosting modules against three baseline methods CLIP, IRRA and RDE. In fairness, when we combine our boosting module with respective baselines, we follow the training strategies and implementation details followed by these baselines. Our boosting weight is set to 1.6 for $k=2$. The weights are updated after every fourth epoch. Furthermore, we observe that when we augment $\mathcal{R}_k$ with pairs for which the correct match occurs at rank-1, the results are better. In Table \ref{comparison} we use `$*$' to denote that the results are computed while boosting samples from $\mathcal{R}_k$. We do not use `$*$' for results obtained from augmented $\mathcal{R}_k$.
 
\subsection{Results} 
\begin{table*}[!h]
\centering
\caption{Comparison with SOTA methods. `+B$^*$'  denotes that weak positives belonging to set $\mathcal{R}_k$ are boosted. `+B' indicates that augmented $\mathcal{R}_k$ set is used for boosting. \textbf{Bold} indicates the best results.}
\resizebox{\textwidth}{!}{
\begin{tabular}{|l|l|cccc|cccc|cccc|cccc|}
\hline
\multirow{2}{*}{\textbf{Methods}} & \multirow{2}{*}{\textbf{Venue}} & \multicolumn{4}{c|}{\textbf{CUHK-PEDES}} & \multicolumn{4}{c|}{\textbf{ICFG-PEDES}} & \multicolumn{4}{c|}{\textbf{RSTPReid}} & \multicolumn{4}{c|}{\textbf{IIITD 20K}} \\ \cline{3-18} 
 &  & R1 & R5 & R10 & mAP & R1 & R5 & R10 & mAP & R1 & R5 & R10 & mAP & R1 & R5 & R10 & mAP \\ \hline
ViTAA \cite{b3} & ECCV'20 & 55.97 & 75.84 & 83.52 & 51.60 & 50.98 & 68.79 & 75.78 & - & - & - & - & - & - & - & - & - \\ 
NAFS \cite{b4} & arXiv'21 & 59.36 & 79.13 & 86.00 & 54.07 & - & - & - & - & - & - & - & - & - & - & - & - \\ 
DSSL \cite{b5} & MM'21 & 59.98 & 80.41 & 87.56 & - & - & - & - & - & 39.05 & 62.60 & 73.95 & - & - & - & - & - \\ 
SSAN \cite{b6} & arXiv'21 & 61.37 & 80.15 & 86.73 & - & 54.23 & 72.63 & 79.53 & - & 43.50 & 67.80 & 77.15 & - & 57.56 & - & - & - \\ 
SAF \cite{b7} & ICASSP'22 & 64.13 & 82.62 & 88.40 & - & - & - & - & - & - & - & - & - & - & - & - & - \\ 
TIPCB \cite{b8} & Neuro'22 & 64.26 & 83.19 & 89.10 & - & - & - & - & - & - & - & - & - & - & - & - & - \\ 
AXM-Net \cite{b9} & MM'22 & 64.44 & 80.52 & 86.77 & 58.73 & - & - & - & - & - & - & - & - & - & - & - & - \\ 
LGUR \cite{b10} & MM'22 & 65.25 & 83.12 & 89.00 & - & 59.02 & 75.32 & 81.56 & - & 45.55 & 68.20 & 77.85 & - & 65.96 & - & - & - \\ 
CFine \cite{b2} & arXiv'22 & 69.57 & 85.93 & 91.15 & - & 60.83 & 76.55 & 82.42 & - & 50.55 & 72.50 & 81.60 & - & - & - & - & - \\

 \hline
 CLIP\cite{b12} & Baseline & 68.55 & 86.32 & 91.85 & 61.25 & 56.80 & 75.99 & 82.31 & 31.93 & 56.00 & 80.85 & 87.90 & 44.06 & 73.48 & 91.82 & 95.6 & 81.69 \\
{CLIP+B}$^*$ & Ours & 70.92 & 88.29 & 93.36 & 63.71 & 60.37 & 78.48 & 84.41 & 34.63 & 56.70 & 80.25 & 87.60 & 44.85 & 74.10 & 92.32  & 95.76 & 82.13 \\
{CLIP+B }  & Ours & 71.44 & 88.65 & 93.29 & 64.24 & 60.79 & 78.61 & 84.59 & 35.02 & 57.00 & 79.60 & 86.80 & 44.47 & 74.80 & 92.66  & 95.98 & 82.70 \\ \hline
IRRA \cite{b1} & CVPR'23 & 73.38 & 89.93 & 93.71 & 66.13 & 63.46 & 80.25 & 85.82 & 38.06 & 60.20 & 81.30 & 88.20 & 47.17 & 77.08 & - & - & - \\ 

{IRRA+B}$^*$& Ours & 73.38 & 89.52 & 93.42 & 66.11 & 63.43 & 80.31 & 85.76 & 38.18 & 59.70 & 80.75 & 87.45 & 46.86 & 77.52 & 93.66 & 96.54 & 84.69 \\
{IRRA+B}& Ours & 74.11 & 89.25 & 93.60 & 66.68 & 63.74 & 80.34 & 85.81 & 38.79 & 60.75 & 82.30 & 90.35 & 47.40 & 78.04 & 93.60 & 96.88 & 85.12 \\ \hline
RDE \cite{b16} & CVPR'24 & 75.94 & 90.14 & 94.12 & 67.56 & \textbf{67.68} & \textbf{82.47} & \textbf{87.36} & \textbf{40.06} & 65.35 & 83.95 & \textbf{89.90} & \textbf{50.88} & 80.90 & 94.48 & 97.02 & 87.04 \\ 
{RDE+B}$^*$ & Ours & \textbf{76.12}   &\textbf{90.29}  &\textbf{94.20}  & \textbf{67.92}  &67.46  &82.28 &87.37&39.95 & 64.40& 83.50& 89.50 & 50.33& \textbf{81.48} &\textbf{94.92} & \textbf{97.30}& \textbf{87.52} \\ 
{RDE+B} & Ours & \textbf{76.12}   &90.12 &93.71  & 67.78  & 67.27  & 82.34 &87.56  & 40.46  & \textbf{65.55} & \textbf{83.80} & 89.35 & 49.47 & 80.80 & 94.56 & 97.12 & 86.96 \\ \hline
\end{tabular}
\label{comparison}
}
\end{table*}
We evaluate the performance of our boosting approach by integrating it with CLIP and state-of-the-art methods such as IRRA and RDE. In Table~\ref{comparison}, we present a comprehensive comparison using our boosted models: CLIP+B, IRRA+B, and RDE+B, against other methods. The baseline results of the vanilla CLIP, IRRA, and RDE models are also included to ensure a fair comparison. 

CLIP+B scores 71.44\% on CUHK-PEDES. It achieves a 2.89\% improvement in rank-1 accuracy when compared against baseline CLIP \cite{b12}. On ICFG-PEDES, CLIP+B shows a 4\% rank-1 improvement over CLIP. We see similar performance gain in mAP also. On IIITD-20K, CLIP+B attains 74.80\% while CLIP scores 73.48\%.

Similarly, IRRA+B demonstrates better performance than IRRA across all datasets in terms of R1 and mAP. RDE+B achieves R1 accuracy of 76.12\% on CUHK-PEDES and 65.55\% on RSTPReid, and outperforms all other methods.
These results demonstrate that our boosting method consistently outperforms others across all four datasets. 
 Notably, all performance metrics across all datasets have shown consistent improvements with our boosted models, further validating the effectiveness of our approach.

\subsection{Evaluation Under Distractors} To assess the robustness of our model, we use distractor images from other pedestrian dataset's test set combined with the training dataset’s test set. The results are reported in Table \ref{tab:cross_dataset_performance}. Our model scores significantly better than CLIP for both CUHK-PEDES and ICFG-PEDES. Similarly, our model gives better rank-1 accuracy compared to IRRA for all three datasets. We note that there is a slight dip in R1 accuracy in case of CLIP+B evaluated on RSTPReid. This may be due to the fact that it is a noisy dataset and in presence of distractors, the retrieval task becomes much more complex.

\begin{table}[ht!]
    \centering
    \caption{Performance comparison under the presence of distractors. Base refers to CLIP or IRRA, and Boost refers to the respective boosted version.}
    \footnotesize
    \setlength{\tabcolsep}{2.4pt}
    
    \begin{tabular}{|l|l|l|l l|c c|c c|}
        \hline
        \multirow{2}{*}{\begin{tabular}[c]{@{}l@{}}Method\end{tabular}} & 
        \multirow{2}{*}{\begin{tabular}[c]{@{}l@{}}Training\\Dataset\end{tabular}} & 
        \multirow{2}{*}{\begin{tabular}[c]{@{}l@{}}Distractor\\Datasets\end{tabular}} & 
        \multicolumn{2}{c|}{R1} & 
        \multicolumn{2}{c|}{R5} & 
        \multicolumn{2}{c|}{R10} \\
        \cline{4-9}
        & & & Base & Boost & Base & Boost & Base & Boost \\
        \hline
        
        \multirow{3}{*}{\begin{tabular}[c]{@{}l@{}}CLIP\end{tabular}}
        & CUHK & ICFG, RST & 61.06 & \textbf{63.21} & 77.81 & \textbf{79.91} & 83.37 &  \textbf{85.43} \\
        & ICFG & CUHK, RST &55.15 &  \textbf{59.27} & 75.02 & \textbf{77.8} & 81.50 & \textbf{83.93} \\
        & RST & CUHK, ICFG & \textbf{24.45} & 24.30 & \textbf{42.40} &41.20 & \textbf{52.65} & 50.20 \\
        \hline
        \multirow{3}{*}{IRRA} 
        & CUHK & ICFG, RST & 65.29 & \textbf{65.97} &  \textbf{81.97} &81.17 & \textbf{86.40} & 85.95 \\
        & ICFG & CUHK, RST & 62.03 & \textbf{62.20} & 79.51 & \textbf{79.61} & \textbf{85.18} & 85.14 \\
        & RST & CUHK, ICFG & 25.20 & \textbf{25.30} & 41.65 & \textbf{44.25} & 49.20 & \textbf{52.65} \\
        \hline
    \end{tabular}
    \label{tab:cross_dataset_performance}
\end{table}

\subsection{Cross-dataset Testing}
We present the cross-dataset testing results in Table \ref{tab:cross_dataset_evaluation}. Our method, CLIP+B, gives significant gains in terms of rank-1 and mAP when trained on CUHK-PEDES or ICFG-PEDES. In the case of training on RSTPReid, we obtain superior results when tested on ICFG-PEDES. 

When we train IRRA+B on CUHK-PEDES and test on ICFG-PEDES or RSTPReid, we see superior performance in terms of mAP. Further, training on ICFG-PEDES and testing on other datasets also outperforms IRRA. When we train on RSTPReid and test on ICFG-PEDES, we see that our method performs better in terms of mAP. However, when we test on CUHK-PEDES, the performance of our model deteriorates due to noisy nature of RSTPReid.
\begin{table}
   \centering
   \caption{Cross-dataset evaluation.}
   \begin{tabular}{|l|l|l|c c|c c|}
       \hline
       
       \multirow{2}{*}{\begin{tabular}[c]{@{}l@{}}Method\end{tabular}} & 
       \multirow{2}{*}{\begin{tabular}[c]{@{}l@{}}Train \end{tabular}} & 
       \multirow{2}{*}{\begin{tabular}[c]{@{}l@{}}Test \end{tabular}} & 
       \multicolumn{2}{c|}{R1}  & \multicolumn{2}{c|}{mAP} \\
       \cline{4-7}
       & & & Base & Boost  & Base   & Boost   \\
       \hline
       \multirow{9}{*}{CLIP} & \multirow{3}{*}{CUHK} 
       
        & ICFG & 37.11 & \textbf{42.08} & 19.24 & \textbf{21.45} \\
       & & RSTP & 48.80 & \textbf{52.95} & 36.47 & \textbf{39.24} \\
       \cline{2-7}
       & \multirow{3}{*}{ICFG} & CUHK & 32.07 & \textbf{33.11}  & 29.75 & \textbf{30.03} \\
       & & RSTP & 40.20 & \textbf{43.90} & 31.28 & \textbf{34.44} \\
       \cline{2-7}
       & \multirow{3}{*}{RSTP} & CUHK & \textbf{30.85} & 30.31  & \textbf{28.98} & 28.53 \\
       & & ICFG & 27.03 & \textbf{30.76} & 17.16 & \textbf{19.07} \\
       \hline
       \multirow{9}{*}{IRRA} & \multirow{3}{*}{\begin{tabular}[c]{@{}l@{}}CUHK\end{tabular}} 
      
       &  ICFG & \textbf{42.80} & 42.68 & 21.80 & \textbf{22.52} \\
       & & RSTP & \textbf{53.20} & 52.70  & 39.59 & \textbf{39.62} \\
       \cline{2-7}
       & \multirow{3}{*}{ICFG} & CUHK & 33.43 & \textbf{34.18} & 31.38 & \textbf{31.97} \\
       & & RSTP & 45.10 & \textbf{46.15} & 36.76 & \textbf{37.31} \\
       \cline{2-7}
       & \multirow{3}{*}{RSTP} & CUHK & \textbf{32.72} & 31.48 &  \textbf{30.18} & 29.03 \\
       & & ICFG & 32.37 & \textbf{32.95} & 20.57 & \textbf{20.81} \\
       
       \hline
       
   \end{tabular}
   \label{tab:cross_dataset_evaluation}
\end{table}

\subsection{Ablation Studies}
\begin{figure}[h] 
    \centering
       \includegraphics[width=.96\columnwidth]{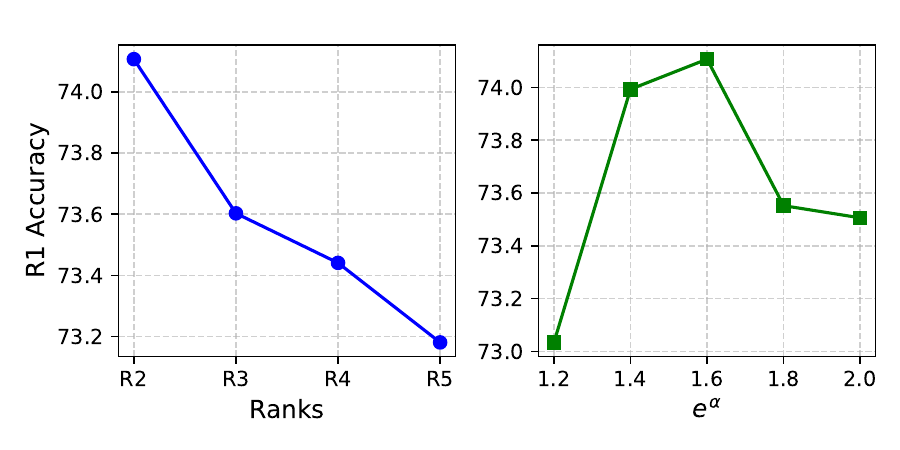}
          \\
       \includegraphics[width=0.56\columnwidth]{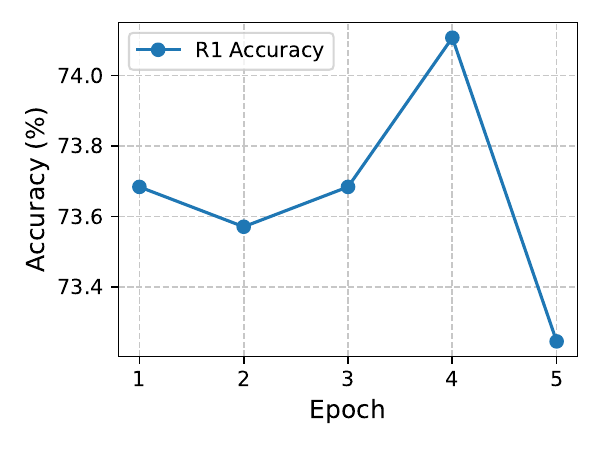}
    \caption{Top: Ablation for $k$ and $exp(\alpha)$. Bottom: Ablation for number of epoch before which weights are updated.}
    \label{fig:ablation}
\end{figure}

We perform the ablation on CUHK-PEDES dataset.
Here, we discuss the results related to the selection of rank-$k$ for which the samples are boosted, values of respective $w^k$ and the number of epochs for which weights are same before being updated.
We evaluated rank-1 accuracy while boosting for different ranks $k= 2, 3, 4, 5$. Fig. \ref{fig:ablation} shows that boosting beyond $k=2$ leads to a drop in the performance. Therefore, we only boost for $k=2$. 

Further, we experimented with various values of $exp(\alpha)$ and selected the one yielding the highest rank-1 accuracy, as shown in Fig. \ref{fig:ablation}. 

We also experiment with different epoch intervals for boosting. Here, we compute the boosting weights and keep it same for a number of consecutive epochs.
In Fig. \ref{fig:ablation} we see that the best results are obtained when we keep the weights same for 4 consecutive epochs. Thus, we re-compute the weights after every 4 epochs. 

\subsection{Qualitative results}
\begin{figure}[ht!]
    \centering
    \includegraphics[width=.925\columnwidth]{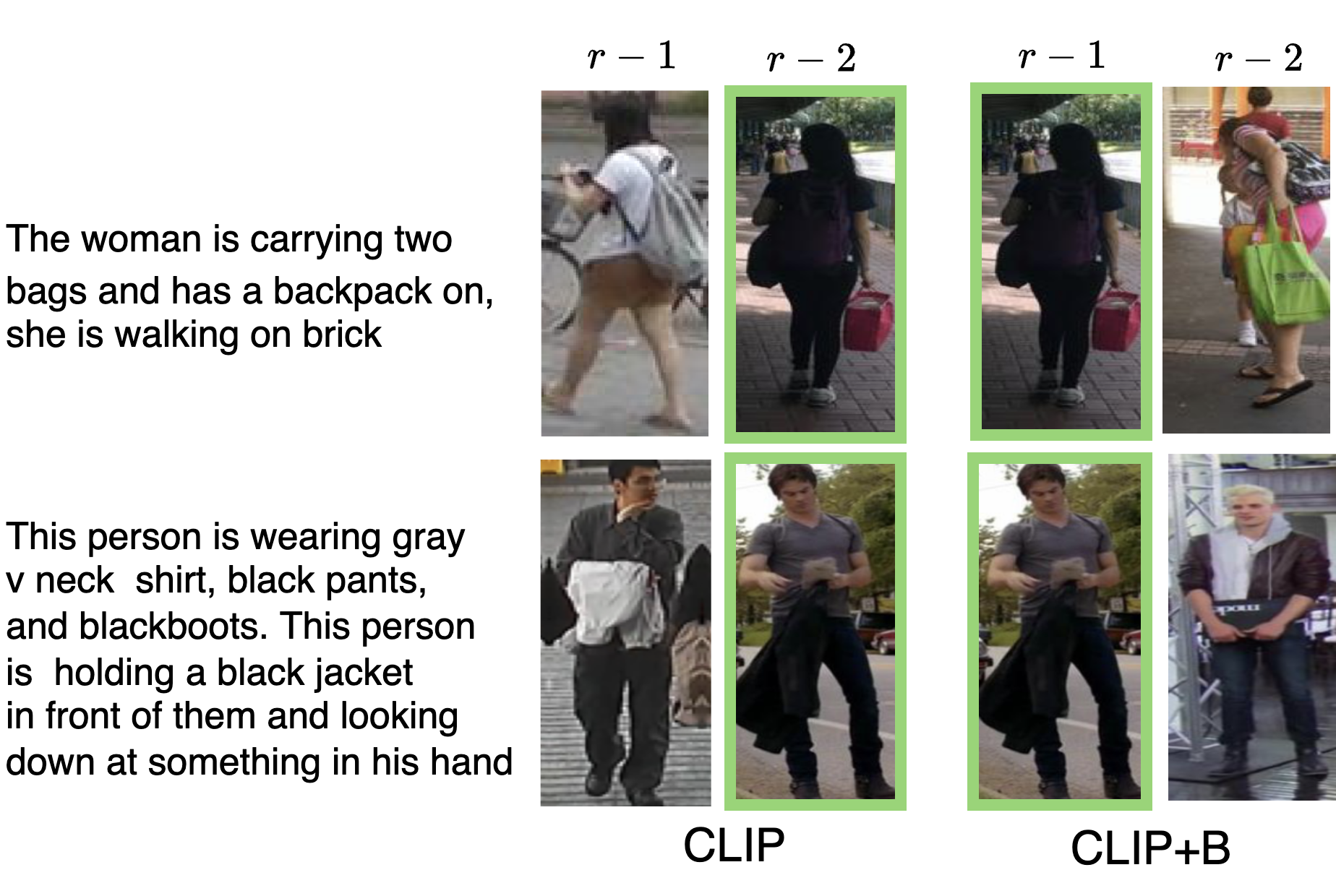}
    \caption{Qualitative results of images retrieved by CLIP and CLIP+B. The green boxes indicate the correct matches. It can be seen that rank-2 correct matches are pushed to rank-1 when retrieved by our boosted model compared to the baseline.}
    \label{fig:retrieval-vis}
\end{figure}

In Fig. \ref{fig:retrieval-vis} we show the qualitative comparison between CLIP and CLIP+B. We show how the rank-2 samples are pushed to rank-1 by our boosted model. Initially the CLIP model failed to retrieve the correct image at rank-1. Instead it chose another slightly similar image. However, our boosting module guided the model to focus on the true match and make the correct retrieval at rank-1.
\section{Conclusion}
In our work, we propose a boosting method and demonstrate its effectiveness by integrating it with various losses. Recognizing the challenges posed by limited and noisy data, we design our system to better utilize difficult samples. We extensively evaluate our model under different settings and compare with the existing methods. Our experiments reveal state-of-the-art performance.
Future work will focus on advancing data filtration techniques and dynamic weighting strategies to further enhance the training.

\end{document}